\pdfoutput=1

\documentclass[11pt]{article}

\usepackage{acl}
\usepackage{graphicx}
\usepackage{times}
\usepackage{latexsym}

\usepackage[T1]{fontenc}

\usepackage[utf8]{inputenc}

\usepackage{microtype}

\title{niksss at HinglishEval: Language-agnostic BERT-based Contextual Embeddings with Catboost for Quality Evaluation of the Low-Resource Synthetically Generated Code-Mixed Hinglish Text}

\author{Nikhil Singh \\
  Manipal University Jaipur \\\
  \texttt{nikhil3198@gmail.com} \\}

\begin{document}
\maketitle
\begin{abstract}
 This paper describes the system description for the HinglishEval challenge at INLG 2022. The goal of this task was to investigate the factors influencing the quality of the code-mixed text generation system. The task was divided into two subtasks, quality rating prediction and annotators’ disagreement prediction of the synthetic Hinglish dataset. We attempted to solve these tasks using sentence-level embeddings, which are obtained from mean pooling the contextualized word embeddings for all input tokens in our text. We experimented with various classifiers on top of the embeddings produced for respective tasks. Our best-performing system ranked 1st on subtask B and 3rd on subtask A. We make our code available here: {\url{https://github.com/nikhilbyte/Hinglish-qEval}}
\end{abstract}

\section{Introduction}
With the increase in popularity of social media platforms like blogs, Facebook, and Twitter in India, the amount of spoken and written Hinglish data has been on the rise. Hinglish is a blend of English and Hindi, involving code-switching between the above-mentioned languages. Due to the increasing number of users, the analysis of this new hybrid language using computational techniques has gotten important in a number of natural language processing applications like machine translation (MT) and speech-to-speech translation. \cite{bali-etal-2014-borrowing},\cite{Das2013CodeMixingIS}.

Classical NLP problems such as language modeling \cite{pratapa-etal-2018-language}, sentiment analysis \cite{singh-lefever-2020-sentiment}, \cite{chakravarthi-etal-2021-findings-shared},Hate-Speech Identification \cite{SREELAKSHMI2020737} and language identification \cite{molina-etal-2016-overview} are covered for Code-Mixed textual data.
However, the generation and evaluation aspect of CM data hasn't been explored a lot. 

This shared task aims to further the research of quality evaluation of the generated code-mixed text in a new way, proposing two tasks that will help quantify the quality of the synthetically generated CM text. Moreover, the organizers put forward another task that will help estimate the disagreement between the different human annotators, which further strengthens and reduce the noisiness of the ground-truth \emph{quality} labels of the generated CM text sequence.

\section{Related Work}
There has been an increased interest in Code-Mixed data for various NLG tasks.\cite{yang2020code} proposed a new pre-training strategy to tackle the complexities in CM text sequences in a non-traditional way. \cite{Gautam2021CoMeTTC} talks about generating low-resource Code-Mixed language from a high resource language such as English using various Seq2Seq models such as mBART \cite{liu2020multilingual}. Other than this, various augmentation techniques were also proposed to improve the quality of generated Hinglish text sequences \cite{50348}. Due to its high linguistic diversity and lack of standardization, the basic Natural Language generation needs to be tackled and evaluated differently as shown in \cite{garg-etal-2021-mipe} where they propose different metrics to evaluate the quality of generated CM data and show why traditional translation metrics such as BLUE \cite{papineni-etal-2002-bleu} etc. cannot capture the quality evaluation properly. 

\section{Task Overview and Dataset}

The task \cite{srivastava-singh-2021-quality} was divided into two subtasks. Subtask-A comprised of predicting the quality of the generated Hinglish sentences
text on a scale of 1–10. 1 is low quality and 10 is the highest quality, considering the semantics and meaningfulness of the generated text sequence. However, the code-mixed language is seldom used in a formal setting, leading the popular evaluation techniques such as BLUE and WER being inappropriate. The organizers tried to tackle this using another way of evaluation to curb the noisiness of labels occurring in subtask-A by proposing another subtask-B. This subtask tests the capacity of the proposed models for estimating the disagreement between individual annotators, which often occurs when trying to evaluate the quality of informal text sequences.

\begin{figure*}[ht]
    \centering
    \includegraphics[width = \textwidth]{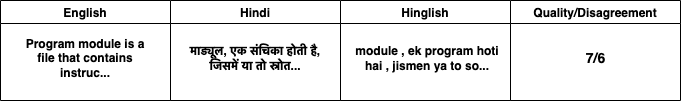}
    \caption{A Single Instance from the Dataset}
    \label{fig:Figure 1}
\end{figure*}

The data for this task introduced in \cite{srivastava2021hinge} is called the HinGE dataset. Its dataset comprises 3,952 instances. Where a particular instance i comprises a text sequence triplet in English, Hindi, and hinglish language and \emph{Average rating} as the label for subtask-A and \emph{Annotator disagreement} as the label for subtask-B. These instances were shuffled and divided into three parts in a ratio of 70:10:20, leading to 2766, 395, and 791 data instances in train, validation, and test respectively. An instance of the dataset can be found in Figure \ref{fig:Figure 1}.

\section{Methodology}
\begin{figure}
    \centering
    \includegraphics[width = \linewidth]{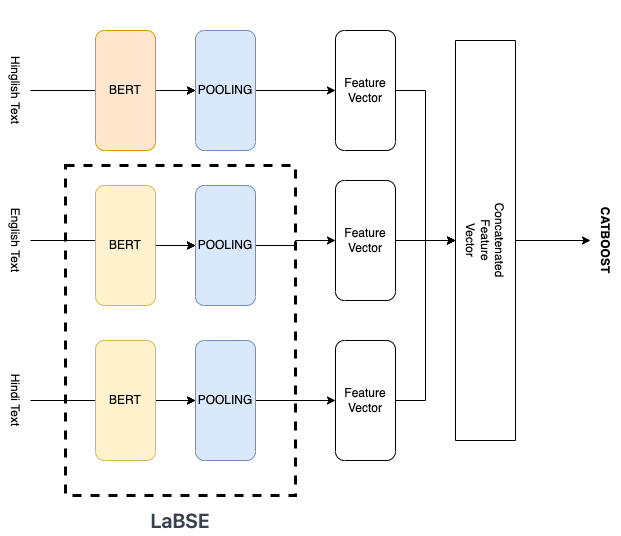}
    \caption{System Design}
    \label{fig:Figure 2}
\end{figure}

We attempted these tasks as a text triplet classification problem, wherein we have three text sequences side-by-side and a label attached to them. We analyzed the text sequences and found them to be clean and without any redundant information, hence we didn't perform any traditional pre-processing step.
The following steps were taken to build the submitend system:
\begin{itemize}
    \item{Out of the three text sequences in a particular data instance, we feed the  English and Hindi input sentences or texts into a transformer network named Language-agnostic BERT sentence embedding model (LaBSE) \cite{feng2020language}. The model produces contextualized word embeddings for all input tokens in our text into a shared latent space that produces similar vector/embeddings for similar sentences in a language-agnostic way. As we want a fixed-sized output representation (vector u), we need a pooling layer. Different pooling options are available, the most basic one is mean-pooling: We simply average all contextualized word embeddings the model is giving us. This gives us a fixed 768-dimensional output vector independent of how long our input text was.}
    \item{The hinglish sequence was embedded using a BERT \cite{devlin2018bert} based model for hinglish text sequences available here \footnote{\url{https://huggingface.co/niksss/Hinglish-HATEBERT}} after using the same strategy as done for the English and Hindi counterparts.}
    \item{By this point, we have the three sentences/texts mapped to a fixed sized dense vector.}
    \item{The obtained vectors are then concatenated and fed into a catboost \cite{prokhorenkova2018catboost} based classifier. }
    
    \item{The model was trained in a supervised manner using the default catboost classifiers with a logloss objective.A seed value of 42 was used to keep the model deterministic.}
    
    \item{The model took approximately 1.75 hours to train on CPU with a memory of 12Gb. }
    
    \item{The complete experiment was done on Google Colab Pro.}
    
    \item{The model architecture can be seen in Figure  \ref{fig:Figure 2}.}
\end{itemize}
All our experiments were performed using SBERT \footnote{\url{https://www.sbert.net/index.html}}

\section{Results}
Three evaluation metrics F1-score (FS),Cohen’s Kappa (CK),Mean Squared Error (MSE) were used to measure the performance of the submitted systems. We present the results obtained on test set along with the baselines in Table  \ref{tab:Table 1} .

\begin{table}
\begin{tabular}{|l|r|c|c|}
\hline
\textbf{SubTask} & \textbf{FS} & \textbf{CK} & \textbf{MSE}\\
\hline
SubTask A & 0.25062 & 0.08153& 2.00000 \\
SubTask B & 0.26115 & - & 3.00000 \\
Baseline A & 0.26637 & 0.09922 & 2.00000 \\
Baseline B & 0.14323 & - & 5.00000 \\\hline
\end{tabular}
\caption{Results on for Test Set}
\label{tab:Table 1}
\end{table}

\section{Conclusion}

We developed a system to evaluate the quality of machine-generated text sequences using a combination of deep learning feature vectors and machine learning models. The results are nowhere near what would actually be used to evaluate the quality of the generated sequence. However, this is the first installment of the shared task and it sets off the baselines for future research on the same subject. 

\bibliography{anthology,custom}
\bibliographystyle{acl_natbib}

\end{document}